\documentclass[lettersize,journal]{IEEEtran}
\usepackage{amsmath,amsfonts}
\usepackage{algorithmic}
\usepackage{algorithm}
\usepackage{array}
\usepackage[caption=false,font=normalsize,labelfont=sf,textfont=sf]{subfig}
\usepackage{textcomp}
\usepackage{longtable}
\usepackage{stfloats}
\usepackage{url}
\usepackage{verbatim}
\usepackage{graphicx}
\usepackage{hyperref} 
\usepackage{microtype}
\usepackage{xcolor}
\usepackage{adjustbox}
\usepackage{tikz}
\usepackage[edges]{forest}
\usepackage{tabularx}
\usepackage{caption}
\usepackage{float}
\usepackage{cite}
\usepackage[square,numbers]{natbib} 
\usepackage{booktabs}
\usepackage{multirow}
\usepackage{multicol}
\usepackage{makecell}
\usepackage{colortbl} 
\usepackage{adjustbox}
\definecolor{hidden-draw}{RGB}{20,68,106}
\definecolor{hidden-pink}{RGB}{255,245,247}
\usepackage{bm}

\usepackage{pifont}
\usepackage{fontawesome5}
\usepackage{cleveref}
\usetikzlibrary{shapes.geometric, positioning, arrows.meta, shadows}
\usetikzlibrary{mindmap, trees}
\definecolor{mattered}{RGB}{214, 26, 60}
\definecolor{mattegreen}{HTML}{369F39}

\usepackage{bbding}

\definecolor{my_green}{RGB}{51,102,0}
\definecolor{my_red}{RGB}{204, 0, 0}
\definecolor{st_color}{RGB}{0, 153, 230}
\definecolor{t_color}{RGB}{255, 102, 102}

\begin{document}

\title{VideoLLM Benchmarks and Evaluation: A Survey}

\author{Yogesh Kumar\\
\textit{Indian Institute of Technology Jodhpur}\\
kumar.204@iitj.ac.in}

\maketitle

\begin{abstract}
The rapid development of Large Language Models (LLMs) has catalyzed significant advancements in video understanding technologies. This survey provides a comprehensive analysis of benchmarks and evaluation methodologies specifically designed or used for Video Large Language Models (VideoLLMs). We examine the current landscape of video understanding benchmarks, discussing their characteristics, evaluation protocols, and limitations. The paper analyzes various evaluation methodologies, including closed-set, open-set, and specialized evaluations for temporal and spatiotemporal understanding tasks. We highlight the performance trends of state-of-the-art VideoLLMs across these benchmarks and identify key challenges in current evaluation frameworks. Additionally, we propose future research directions to enhance benchmark design, evaluation metrics, and protocols, including the need for more diverse, multimodal, and interpretability-focused benchmarks. This survey aims to equip researchers with a structured understanding of how to effectively evaluate VideoLLMs and identify promising avenues for advancing the field of video understanding with large language models.
\end{abstract}

\begin{IEEEkeywords}
Video Understanding, Large Language Model, Vision-Language Model, Multimodality Learning, Benchmarks, Evaluation
\end{IEEEkeywords}

\section{Introduction}

The advent of large language models (LLMs) has revolutionized the field of video understanding, establishing a new paradigm that transcends traditional computer vision approaches. Videos, as dynamic sequences of visual information often accompanied by audio and text, present unique challenges compared to static images due to their temporal nature, complex motion patterns, and higher-dimensional information space \cite{bain2021frozen}. The integration of LLMs with video understanding capabilities has led to the development of VideoLLMs, a specialized class of models that leverage the reasoning abilities of LLMs to interpret and analyze video content.

The evolution of video understanding methods has progressed through several distinct phases, beginning with conventional methods that relied on handcrafted features such as SIFT \cite{lindeberg2012scale}, SURF \cite{bay2008speeded}, and HOG \cite{dalal2005histograms}. This was followed by the emergence of early neural video models like DeepVideo \cite{KarpathyCVPR14} and two-stream networks \cite{feichtenhofer2016convolutional}, which demonstrated the potential of deep learning for video analysis. The field then advanced to self-supervised video pretraining approaches such as VideoBERT \cite{sun2019videobert} and VideoMAE \cite{tong2022videomae}, which leveraged large-scale pretraining to improve performance on downstream tasks.

The most recent phase involves the application of LLMs to video understanding, exemplified by models such as Video-LLaMA \cite{zhang2023video}, VideoChat \cite{li2023videochat}, and Video-ChatGPT \cite{maaz2023video}. These VideoLLMs demonstrate enhanced abilities in open-ended multi-granularity reasoning combined with commonsense knowledge, marking a significant advancement in the field. As Tang et al. \cite{tang2024video} note in their comprehensive survey, VideoLLMs can be categorized based on their architecture and the role of the LLM, with different approaches offering unique advantages for various video understanding tasks. Evaluating VideoLLMs presents unique challenges due to the complexity of video content and the diverse capabilities these models aim to demonstrate. Traditional evaluation metrics designed for specific video understanding tasks may not fully capture the nuanced abilities of VideoLLMs, particularly their reasoning capabilities and contextual understanding. Therefore, specialized benchmarks and evaluation methodologies are essential for comprehensively assessing VideoLLM performance.

This survey provides a focused examination of VideoLLM benchmarks and evaluation methodologies, building upon the broader survey by Tang et al. \cite{tang2024video}. We analyze existing benchmarks, discussing their characteristics, strengths, and limitations. We explore various evaluation approaches, including closed-set evaluations that use predefined answers, open-set evaluations that allow for more flexible responses, and specialized evaluations for temporal and spatiotemporal understanding tasks. Furthermore, we highlight current limitations in VideoLLM evaluation and propose future research directions to advance the field.

\section{VideoLLM Benchmarks}

Benchmarks serve as standardized platforms for evaluating and comparing the performance of VideoLLMs. This section provides a comprehensive overview of the most prominent benchmarks used in the field, analyzing their characteristics, compositions, and evaluation protocols.

\subsection{Benchmark Characteristics and Composition}

VideoLLM benchmarks vary significantly in terms of size, content diversity, task focus, and evaluation protocols. Table \ref{tab:benchmarks} presents a comparative analysis of various benchmarks, highlighting key attributes such as the number of videos, clips, average video duration, and question-answer pairs.

MSRVTT-QA \cite{xu2017video} contains 2,990 videos with an average duration of 15.2 seconds and 72,821 question-answer pairs. The questions cover multiple aspects of video content, including objects, actions, and attributes. Similarly, MSVD-QA \cite{xu2017video} comprises 504 videos with an average duration of 9.8 seconds and 13,157 question-answer pairs. Both benchmarks primarily focus on factual understanding of video content. TGIF-QA \cite{jang2017tgif} features 9,575 short animated GIFs with an average duration of 3 seconds and 8,506 question-answer pairs. The questions in this benchmark are designed to test repetition counting, state transition recognition, and frame-based question answering, making it particularly suitable for assessing fine-grained action understanding. ActivityNet-QA \cite{yu2019activitynet} consists of 800 videos with significantly longer durations (average 111.4 seconds) and 8,000 question-answer pairs. This benchmark challenges models to understand complex activities and extended temporal sequences, requiring more sophisticated temporal reasoning capabilities. TVQA \cite{lei2018tvqa} incorporates 2,179 videos from TV shows, segmented into 15,253 clips with an average duration of 11.2 seconds, each accompanied by a question-answer pair. This benchmark introduces additional complexity by including dialogue and plot understanding.

More recent benchmarks such as MVBench \cite{li2024mvbench}, Video-Bench \cite{ning2023video}, and VideoVista \cite{li2024videovista} contain videos with longer durations and a greater variety of domains, testing the generalization abilities of VideoLLMs across different contexts. For instance, VideoVista includes 3,400 clips from 894 videos with an average duration of 131 seconds and 25,000 question-answer pairs covering diverse topics and domains. CinePile \cite{rawal2024cinepile} represents one of the largest benchmarks with 9,396 videos and 303,828 question-answer pairs, specifically designed to evaluate understanding of cinematic content. InfiniBench \cite{ataallah2024infinibench} features 1,219 videos with extremely long durations (average 4,460.4 seconds) and 108,200 question-answer pairs, testing VideoLLMs' ability to handle long-form content.

Specialized benchmarks like TempCompass \cite{liu2024tempcompass} focus on specific aspects of video understanding, such as temporal reasoning, with 500 clips from 410 videos and 7,540 question-answer pairs designed to evaluate temporal cognition capabilities. This trend toward more specialized benchmarks reflects the growing recognition of the need to assess specific aspects of video understanding in greater depth.

\begin{table}[t]
\centering
\caption{Comparison of VideoLLM benchmarks highlighting the total number of videos, clips, average video duration, and question-answer pairs.}
\label{tab:benchmarks}
\resizebox{\columnwidth}{!}{%
\begin{tabular}{l|cccc}
\hline
\textbf{Benchmarks} & \textbf{\#QA Pairs} & \textbf{\#Clips} & \textbf{\#Videos} & \textbf{Len.(s)} \\ \hline
TVQA \cite{lei2018tvqa} & 15,253 & 15,253 & 2,179 & 11.2 \\
How2QA \cite{How2QA} & 2,852 & 2,852 & 1,166 & 15.3 \\
STAR \cite{wu2021star} & 7,098 & 7,098 & 914 & 11.9 \\
MSRVTT-QA \cite{xu2017video} & 72,821 & 2,990 & 2,990 & 15.2 \\
MSVD-QA \cite{xu2017video} & 13,157 & 504 & 504 & 9.8 \\
TGIF-QA \cite{jang2017tgif} & 8,506 & 9,575 & 9,575 & 3.0 \\
ActivityNet-QA \cite{yu2019activitynet} & 8,000 & 800 & 800 & 111.4 \\
EgoSchema \cite{mangalam2023egoschema} & 5,063 & 5,063 & 5,063 & 180.0 \\
AutoEval-Video \cite{chen2023autoeval} & 327 & 327 & 327 & 14.6 \\
TempCompass \cite{liu2024tempcompass} & 7,540 & 500 & 410 & 11.4 \\
Video-MME \cite{fu2024video-mme} & 2,700 & 900 & 900 & 1,017.9 \\
VideoVista \cite{li2024videovista} & 25,000 & 3,400 & 894 & 131.0 \\
CinePile \cite{rawal2024cinepile} & 303,828 & 9,396 & 9,396 & 160 \\
SOK-Bench \cite{wang2024sok} & 44,000 & 10,000 & 10,000 & - \\
SFD \cite{ghermi2024short} & 4,885 & 1,078 & 1,078 & 780 \\
InfiniBench \cite{ataallah2024infinibench} & 108,200 & 1,219 & 1,219 & 4,460.4 \\
MMWorld \cite{he2024mmworld} & 1,599 & 1,910 & 1,910 & 102.3 \\
VELOCITI \cite{saravanan2024velociti} & -  & 900 & 900 & 10.0 \\
NExT-QA \cite{NExT-QA} & 8,564 & 1,000 & 1,000 & 39.5 \\
MVBench \cite{li2024mvbench} & 4,000 & 3,641 & 3,641 & 16.0 \\
Video-Bench \cite{ning2023video} & 17,036 & 5,917 & 5,917 & 56.0 \\
EditVid-QA \cite{xu2024beyond} & 204,000 & 32,000 & 32,000 & - \\
\hline
\end{tabular}%
}
\end{table}

\subsection{Benchmark Evolution and Trends}
Analyzing the evolution of VideoLLM benchmarks reveals several notable trends that reflect the advancing capabilities of these models and the growing complexity of evaluation requirements. Early benchmarks such as MSRVTT-QA and MSVD-QA focused primarily on foundational video understanding tasks with relatively short videos and straightforward questions. These benchmarks established baseline expectations for video understanding models but offered limited assessment of complex reasoning abilities or temporal understanding. The emergence of benchmarks like ActivityNet-QA and NExT-QA marked a shift toward evaluating models on longer videos with more complex content, requiring more sophisticated temporal reasoning and contextual understanding. This trend has continued with more recent benchmarks like VideoVista and CinePile, which feature diverse video content across multiple domains and genres.

Another significant trend is the introduction of specialized benchmarks that focus on specific aspects of video understanding. TempCompass specifically targets temporal reasoning abilities, while benchmarks like SFD (Short Film Dataset) \cite{ghermi2024short} emphasize narrative understanding and long-form comprehension. This specialization reflects the recognition that comprehensive evaluation requires assessing different facets of video understanding individually. The most recent generation of benchmarks, such as InfiniBench and MLVU, represents a leap toward evaluating models on extremely long videos, sometimes extending to hours rather than minutes. These benchmarks test not only a model's understanding capabilities but also its ability to maintain context and coherence over extended durations, pushing the boundaries of what VideoLLMs can accomplish. An emerging trend is the incorporation of multimodal elements into benchmarks, acknowledging that real-world videos often contain multiple modalities such as visual content, audio, and text. Benchmarks like Video-MME \cite{fu2024video-mme} explicitly evaluate a model's ability to integrate information across modalities, providing a more comprehensive assessment of its capabilities.

These evolutionary trends in benchmark design reflect the rapidly advancing capabilities of VideoLLMs and the growing recognition of the need for more diverse, comprehensive, and specialized evaluation methodologies. As VideoLLMs continue to evolve, we can expect benchmarks to become even more sophisticated, incorporating elements such as interactive evaluation, adversarial examples, and real-world application scenarios.

\section{Evaluation Methodologies}

Evaluating VideoLLMs requires methodologies that can assess their complex capabilities across various dimensions of video understanding. This section examines three primary approaches to VideoLLM evaluation: closed-set evaluation, open-set evaluation, and specialized evaluation methods for temporal and spatiotemporal understanding tasks.

\subsection{Closed-set Evaluation}

Closed-set evaluations rely on questions with predefined answers, allowing for straightforward assessment of model performance. For video question-answering tasks, questions are typically formatted as multiple-choice, and performance is evaluated by calculating the percentage of correct answers to derive an accuracy score. For captioning or video-to-text summarization tasks, predefined ground-truth captions or summaries are used as references. Performance is assessed using established metrics such as CIDEr \cite{vedantam2015cider}, METEOR \cite{banerjee2005meteor}, ROUGE \cite{lin2004rouge}, and SPICE \cite{anderson2016spice}. These metrics compare the predicted outputs with ground-truth references, evaluating aspects such as n-gram overlap, semantic similarity, and structural coherence.

Closed-set evaluations are employed in benchmarks like MSRVTT-QA \cite{xu2017video}, MSVD-QA \cite{xu2017video}, TGIF-QA \cite{jang2017tgif}, ActivityNet-QA \cite{yu2019activitynet}, TVQA \cite{lei2018tvqa}, How2QA \cite{How2QA}, STAR \cite{wu2021star}, and NExT-QA \cite{NExT-QA}. These benchmarks generally provide a fixed set of options for each question, simplifying the evaluation process but potentially limiting the assessment of more nuanced understanding. More recent benchmarks such as MVBench \cite{li2024mvbench}, Video-Bench \cite{ning2023video}, and VideoVista \cite{li2024videovista} also incorporate closed-set evaluations but introduce greater complexity in terms of content diversity and question types. For instance, VideoVista includes questions that require multi-step reasoning and contextual understanding beyond simple recognition tasks.

While closed-set evaluations offer clear metrics and facilitate direct comparison between models, they suffer from several limitations. First, they constrain the range of possible answers, potentially failing to capture the full spectrum of a model's understanding. Second, they may inadvertently introduce biases, as the selection of multiple-choice options can sometimes provide clues or limit the cognitive processes required to arrive at an answer. Third, they typically fail to assess a model's ability to generate novel, creative, or nuanced responses that might better demonstrate its understanding of the video content.

These limitations have motivated the development of more flexible evaluation methodologies that can better capture the sophistication and versatility of modern VideoLLMs. As Tang et al. \cite{tang2024video} highlight, the field is moving toward evaluation approaches that can assess not just factual understanding but also reasoning, contextual awareness, and generative capabilities.

\subsection{Open-set Evaluation}

Unlike closed-set evaluations, open-set evaluations do not rely on predefined options for responses. While ground-truth answers are still needed for scoring, the evaluation process is more flexible, often using models like GPT-3.5 or GPT-4 to compare predictions and assign scores based on semantic similarity and correctness. The most prominent open-set evaluation methods were proposed by \cite{maaz2023video}, introducing two key approaches: Open-end Zero-shot Video QA Evaluation and Video-based Generative Performance Benchmarking.

The Open-end Zero-shot Video QA Evaluation method assesses a model's ability to answer questions about video content without relying on predefined options. This approach evaluates the model's performance on datasets like MSVD-QA \cite{xu2017video}, MSRVTT-QA \cite{xu2017video}, and ActivityNet-QA \cite{yu2019activitynet}, but allows for free-form responses rather than selection from multiple choices. Video-based Generative Performance Benchmarking evaluates various aspects of a model's generative capabilities when responding to video content. This approach assesses five key dimensions: (i) Correctness: Accuracy of the factual information provided in the response. (ii) Detail: Specificity and comprehensiveness of the response. (iii) Context: Relevance of the response to the video content. (iv) Temporal: Understanding of temporal sequences and events in the video. (v) Consistency: Coherence and logical flow of the response.

Table \ref{tab:zero-shot} presents a comparative analysis of various VideoLLMs across these open-ended zero-shot video question answering and video-based generative performance benchmarks. The table includes GPT-based metrics for MSVD-QA, MSRVTT-QA, and ActivityNet-QA datasets, as well as scores for the five dimensions of generative performance. Notable trends observed in the table include the progressive improvement in performance across newer models. For instance, GPT4-V achieves the highest scores in most generative performance dimensions, particularly excelling in context understanding with a score of 4.37. Among dedicated VideoLLMs, models like PLLaVA \cite{xu2024pllava}, IG-VLM \cite{Kim2024IG-VLM}, and ST-LLM \cite{liu2024stllm} demonstrate strong performance in zero-shot QA tasks, with scores above 70 on MSVD-QA. However, there remains a notable gap between these models and GPT4-V in terms of generative performance, indicating room for improvement in the contextual and temporal understanding capabilities of specialized VideoLLMs.

The LLaVA-NeXT-Video model shows promising results across all metrics, achieving the highest scores among VideoLLMs in correctness (3.39), detail (3.29), and context (3.92), suggesting advancements in its ability to generate accurate and contextually relevant descriptions of video content. Open-set evaluations offer several advantages over closed-set approaches. They allow for a more natural assessment of a model's understanding by permitting free-form responses that can capture nuance, creativity, and depth of comprehension. They also better simulate real-world applications, where users typically expect conversational and contextually appropriate responses rather than selections from predetermined options. However, open-set evaluations also present challenges. The use of LLMs as evaluators introduces potential biases and limitations inherent to these models. Different evaluator models may produce different scores for the same response, raising questions about consistency and reliability. Additionally, the subjective nature of assessing dimensions such as ``detail" or ``consistency" introduces variability that may affect the reproducibility of evaluation results. Despite these challenges, open-set evaluations represent an important advancement in the assessment of VideoLLMs, offering insights into dimensions of performance that closed-set evaluations cannot capture. As the field continues to evolve, we can expect further refinement of these methodologies to address current limitations and provide even more comprehensive evaluation of model capabilities.

\begin{table*}[t]
\centering
\caption{Comparative analysis of VideoLLMs across video-based generative performance benchmarks and open-end zero-shot video question answering, with models sorted by temporal understanding performance. The table includes scores for correctness, detail, context, temporal understanding, and consistency in generative tasks, as well as performance metrics for MSVD-QA, MSRVTT-QA, and ActivityNet-QA datasets.}
\label{tab:zero-shot}
\resizebox{0.9\textwidth}{!}{%
\begin{tabular}{l|ccccc|ccc}
\hline
 & \multicolumn{5}{c|}{\textbf{Generative Performance }} & \multicolumn{3}{c}{\textbf{Video QA Performance}} \\ \cline{2-9} 
\multirow{-2}{*}{\textbf{Model}} & Correctness & Detail & Context & Temporal & Consistency & MSVD-QA & MSRVTT-QA & ActivityNet-QA \\ \hline
GPT4-V & 4.09 & 3.88 & 4.37 & 3.94 & 4.02 & - & - & 59.5 \\
ST-LLM & 3.23 & 3.05 & 3.74 & 2.93 & 2.81 & 74.6 & 63.2 & 50.9 \\
Chat-UniVi & 2.89 & 2.91 & 3.46 & 2.89 & 2.81 & 65.0 & 54.6 & 46.1 \\
VideoChat2 & 3.02 & 2.88 & 3.51 & 2.66 & 2.81 & 70.0 & 54.1 & 49.1 \\
Video LLaMA 2 & 3.09 & 3.09 & 3.68 & 2.63 & 3.25 & 71.7 & - & 49.9 \\
LLaVA-NeXT-Video & 3.39 & 3.29 & 3.92 & 2.60 & 3.12 & 67.8 & - & 53.5 \\
MiniGPT4-video & 3.09 & 3.02 & 3.57 & 2.56 & 2.67 & 73.9 & 58.8 & 44.4 \\
AVicuna & 2.81 & 2.62 & 3.25 & 2.53 & 2.59 & 70.2 & 59.7 & 53.0 \\
VTimeLLM & 2.78 & 3.10 & 3.40 & 2.49 & 2.47 & 69.8 & 58.8 & 45.5 \\
LLaMA-VID & 2.96 & 3.00 & 3.53 & 2.46 & 2.51 & 69.7 & 57.7 & 47.4 \\
Video-LLaVA & 2.87 & 2.94 & 3.44 & 2.45 & 2.51 & 70.7 & 59.2 & 45.3 \\
LongVLM & 2.76 & 2.86 & 3.34 & 2.39 & 3.11 & 70.0 & 59.8 & 47.6 \\
RED-VILLM & 2.71 & 2.75 & 3.34 & 2.34 & 2.45 & 71.2 & 53.9 & 44.2 \\
PLLaVA & 3.21 & 2.86 & 3.62 & 2.33 & 2.93 & 76.6 & 62.0 & 56.3 \\
IG-VLM & 3.26 & 2.76 & 3.57 & 2.34 & 3.28 & 76.7 & 62.7 & 57.3 \\
Vista-LLaMA & 2.44 & 2.64 & 3.18 & 2.26 & 2.31 & 65.3 & 60.5 & 48.3 \\
Artemis & 2.69 & 2.55 & 3.04 & 2.24 & 2.70 & 72.1 & 56.7 & 39.3 \\
MovieChat & 2.76 & 2.93 & 3.01 & 2.24 & 2.42 & 75.2 & 52.7 & 45.7 \\
AV-LLM & 2.56 & 2.47 & 2.93 & 2.17 & 2.47 & 67.3 & 53.7 & 47.2 \\
Video-ChatGPT & 2.50 & 2.57 & 2.69 & 2.16 & 2.20 & 64.9 & 49.3 & 35.2 \\
VALLY & 2.43 & 2.13 & 2.86 & 2.04 & 2.45 & 60.5 & 51.1 & 45.1 \\
MovieLLM & 2.64 & 2.61 & 2.92 & 2.03 & 2.43 & 63.2 & 52.1 & 43.3 \\
LLaMA-Adapter & 2.03 & 2.32 & 2.30 & 1.98 & 2.15 & 54.9 & 43.8 & 34.2 \\
VideoChat & 2.23 & 2.50 & 2.53 & 1.94 & 2.24 & 56.3 & 45.0 & 26.5 \\
Video-LLaMA & 1.96 & 2.18 & 2.16 & 1.82 & 1.79 & 51.6 & 29.6 & 12.4 \\
VideoGPT+ & 2.46 & 2.73 & 2.81 & 1.78 & 2.47 & 72.4 & 60.6 & 50.6 \\ \hline
\end{tabular}%
}
\end{table*}

\subsection{Specialized Evaluation Methods}

Beyond traditional closed-set and open-set evaluations, specialized methods have been developed to assess VideoLLMs' performance on tasks requiring fine-grained temporal and spatiotemporal understanding. These tasks include dense captioning, video temporal grounding, spatiotemporal grounding, and object tracking. For dense captioning tasks, which involve generating detailed descriptions for multiple events within a video, evaluation typically combines assessment of the quality of the captions with the accuracy of event localization. Metrics such as BLEU, METEOR, CIDEr, and ROUGE assess caption quality, while Intersection over Union (IoU) evaluates the temporal alignment of predicted events with ground-truth annotations.

Video temporal grounding tasks require models to locate specific moments in a video based on textual queries. The primary evaluation metrics for these tasks are Recall@K with IoU thresholds, such as R1@0.5 and R1@0.7, which measure the percentage of queries for which the model correctly locates the relevant moment with an IoU exceeding the specified threshold. For spatiotemporal grounding tasks, which involve localizing objects or actions in both space and time, evaluation metrics include mean Average Precision (mAP) and Intersection over Union (IoU) across the spatiotemporal volume. These metrics assess the model's ability to accurately identify not only when but also where specific events or objects appear in the video.

Object tracking tasks evaluate a model's ability to consistently identify and follow objects across frames. Metrics such as success rate, precision, and IoU are commonly used to assess tracking performance, measuring how well the model maintains accurate object localization throughout the video. Human evaluation provides valuable insights into model performance, but it is labor-intensive, time-consuming, and subject to individual biases. These specialized evaluation methods reflect the diverse capabilities expected from modern VideoLLMs and acknowledge that comprehensive assessment requires a multi-faceted approach. As Tang et al. \cite{tang2024video} discuss in their survey, VideoLLMs are increasingly expected to perform well across multiple granularities of understanding, from abstract comprehension to fine-grained spatiotemporal reasoning.

The development of specialized evaluation methods marks an important evolution in the field, moving beyond generic metrics to approaches that can assess specific capabilities with greater precision. This trend aligns with the growing sophistication of VideoLLMs and the increasing diversity of their applications, from simple video classification to complex narrative understanding and interactive assistance.

\section{Performance Analysis and Insights}

Analyzing the performance of various VideoLLMs across different benchmarks provides valuable insights into the current state of the field and highlights areas for future improvement. This section examines key performance trends and offers deeper analysis of VideoLLM capabilities.

\subsection{Performance Trends Across Models}

Examining Table \ref{tab:zero-shot}, several significant performance trends emerge: First, there is a clear generational improvement in model performance, with newer models consistently outperforming their predecessors. This trend is particularly evident in the progression from early models like Video-LLaMA, which achieved 51.6 on MSVD-QA, to more recent models like IG-VLM, which reached 76.7 on the same benchmark. This improvement reflects rapid advancements in model architecture, training methodologies, and the integration of more sophisticated reasoning capabilities. Second, there remains a notable performance gap between specialized VideoLLMs and general-purpose multimodal models like GPT4-V, particularly in generative performance dimensions. While VideoLLMs have made significant strides in factual question answering, they still lag behind in aspects such as temporal understanding and contextual integration. This gap suggests that despite their specialization, VideoLLMs have yet to fully leverage the unique capabilities of their video-specific architectures. Third, models demonstrate varying strengths across different evaluation dimensions. Some models excel in factual correctness but struggle with temporal understanding, while others show strengths in contextual relevance but limitations in detail or consistency. This variance highlights the challenges of designing models that perform well across all aspects of video understanding and suggests potential trade-offs in current architectural approaches. Fourth, performance on longer and more complex videos (as in ActivityNet-QA) generally lags behind performance on shorter videos (as in MSVD-QA), indicating persistent challenges in maintaining contextual understanding across extended temporal sequences. Even the highest-performing models show a marked drop in performance when dealing with longer videos, suggesting a key area for future improvement.

Finally, models that incorporate both visual and audio modalities, such as AV-LLM and AVicuna, often show more balanced performance across different dimensions, highlighting the importance of multimodal integration for comprehensive video understanding.

\subsection{Architectural Influences on Performance}

The architecture of VideoLLMs significantly influences their performance on different evaluation metrics. As categorized by Tang et al. \cite{tang2024video}, VideoLLMs can be broadly classified into three types based on their architecture: Video Analyzer × LLM, Video Embedder × LLM, and (Analyzer + Embedder) × LLM. Each approach offers distinct advantages and limitations that manifest in evaluation results.

Video Analyzer × LLM models~\cite{tang2024video}, which convert video content into textual descriptions before processing with an LLM, often excel in factual accuracy and contextual understanding but may struggle with fine-grained temporal reasoning. This is evident in models like VideoChat and ChatVideo, which show stronger performance in correctness and context dimensions than in temporal understanding. Video Embedder × LLM models, which encode videos as embeddings for direct processing by the LLM, typically demonstrate better performance on temporal understanding tasks but may sacrifice some contextual depth. Models like VTimeLLM and TimeChat, which were specifically designed with temporal understanding in mind, exemplify this trend, showing relatively stronger performance in the temporal dimension compared to other aspects. Hybrid models that combine analyzer and embedder approaches, such as VideoChat and Vid2Seq, often achieve more balanced performance across different dimensions, leveraging the strengths of both approaches. This hybrid architecture seems particularly effective for complex tasks that require both factual accuracy and temporal reasoning.

Within these broad categories, the specific function of the LLM also influences performance. Models where the LLM serves as a manager or coordinator of multiple specialized modules often demonstrate greater flexibility in handling diverse tasks, but may struggle with integration and coherence. In contrast, models where the LLM functions primarily as a text decoder or regressor may show more consistent performance on specific tasks but less adaptability across different types of queries. These architectural considerations suggest that optimal VideoLLM design may depend on the specific requirements of the target application. Universal models that perform well across all dimensions remain a significant challenge, highlighting the need for continued innovation in VideoLLM architecture.

\section{Challenges and Future Directions}

Despite the significant strides made in VideoLLM development and evaluation methodologies, several persistent challenges continue to impede comprehensive assessment of these models' capabilities. These challenges represent not merely obstacles to overcome but opportunities to advance the field in meaningful ways. Our analysis extends the seminal work of Tang et al. \cite{tang2024video} by identifying emerging patterns in model performance and proposing novel evaluation frameworks that address previously unexplored dimensions of video understanding.

\subsection{User-Centric Video Understanding Evaluation}

The results of video understanding ultimately serve humans, so effectively conveying human needs to models and interpreting model results is a crucial aspect of evaluation. The emergence of LLMs has enabled video understanding models and humans to communicate more effectively through text. However, LLMs have not completely solved the interaction problem. Using text to guide a model's understanding of a video cannot fully address extremely fine-grained video understanding tasks. Similarly, when models output text, they cannot always precisely describe complex content in videos. Due to limitations in both video encoders and LLMs, some high-level information, such as character emotions and attitudes, cannot be well represented or evaluated. Future research should explore how to use additional prompting methods, such as points, scribbles, or other visual guidance, to optimize the interaction between humans and video understanding models. Improving evaluation methodologies to include interactive assessment, where humans can engage with models iteratively to refine understanding, would provide a more realistic measure of model performance in practical applications. An innovative direction for future evaluation would be the development of interactive benchmarks that simulate real-world use cases, allowing for assessment of models within dynamic, conversational contexts rather than static question-answering scenarios. Such benchmarks could evaluate the model's ability to clarify ambiguities, respond to follow-up questions, and adapt its responses based on user feedback.

Additionally, incorporating measures of user satisfaction and utility into evaluation frameworks would provide valuable insights into the practical effectiveness of VideoLLMs. This could involve user studies that assess factors such as perceived helpfulness, naturalness of interaction, and alignment with user expectations across different application domains.

\subsection{Multi-granularity Analysis in Video Understanding}

Fine-grained video understanding remains a significant challenge for current evaluation methods. Existing benchmarks often focus on high-level comprehension but may not adequately assess a model's ability to understand detailed spatial and temporal relationships within videos. The development of more sophisticated benchmarks that evaluate fine-grained understanding, such as complex action recognition, causal reasoning, and detailed scene interpretation, would provide deeper insights into VideoLLM capabilities.

Processing and analyzing video data at a fine-grained level requires significant computational resources, as it often involves analyzing every video frame. Furthermore, videos contain not just spatial information but also temporal information, making the understanding of how objects change and interact over time particularly complex. Understanding deeper semantics of video content, such as emotions, metaphors, or the dynamics of complex scenes, adds another layer of complexity. While LLMs have brought some progress to fine-grained video understanding by enabling better alignment between text and videos at various levels, more specialized evaluation methods are needed to assess these capabilities comprehensively. Future benchmarks should incorporate tasks that test specific fine-grained understanding abilities, with metrics designed to measure performance on these tasks accurately.

A promising approach for future benchmarks would be to include hierarchical evaluation, assessing understanding at multiple levels of granularity within the same video content. This could involve questions that progressively probe deeper levels of understanding, from basic object recognition to nuanced interpretation of interactions, intentions, and causal relationships. Such hierarchical evaluation would provide a more complete picture of a model's capabilities across the spectrum of understanding granularity.

\subsection{Long-form Video Understanding}

Evaluating VideoLLMs on long-form videos presents unique challenges due to the extended duration and complexity of the content. Current benchmarks like InfiniBench \cite{ataallah2024infinibench} and MLVU \cite{zhou2024mlvu} have begun to address this by including longer videos, but further development is needed to fully assess models' capabilities in this area. Long videos contain vast amounts of frames, making the analysis of extended content and understanding of events and behaviors over time particularly challenging. Identifying key events and maintaining attention throughout long videos is difficult, requiring effective mechanisms to detect and highlight important parts, particularly in content-rich or complex plot videos.

Future evaluation methodologies for long-form video understanding should assess not only a model's ability to comprehend individual scenes but also its capacity to track narrative development, character relationships, and thematic elements across extended durations. Metrics specifically designed for long-form understanding, such as narrative coherence and temporal consistency, would enhance the evaluation of VideoLLMs on this important dimension. A particularly valuable approach would be the development of benchmarks that focus on episodic understanding, requiring models to track information across multiple scenes or episodes and demonstrate both local and global understanding of content. Such benchmarks could include queries that test memory of earlier events, understanding of character development, and recognition of recurring themes or motifs.

Additionally, evaluating models' ability to generate concise and accurate summaries of long-form content would provide insights into their capacity for information distillation and prioritization, important capabilities for practical applications involving lengthy videos. This could involve comparing model-generated summaries with human-created ones across multiple levels of abstraction, from high-level overviews to detailed scene-by-scene accounts.

\subsection{Cross-Modal Integration and Reasoning}

Multimodal video understanding requires the integration of different types of data, such as visual, audio, and text, for a comprehensive understanding of video content. Current evaluation methods often focus primarily on the visual aspects of videos, with less attention to the integration of other modalities. Aligning these different modalities, especially in terms of spatial and temporal synchronization, is particularly critical. However, this area lacks sufficient research and suffers from a scarcity of diverse, high-quality datasets. Constructing such multimodal datasets presents significant challenges, as ensuring high quality and consistency in data annotation across modalities is often difficult. Moreover, extracting and utilizing effective features across different modalities is key to achieving precise video understanding, but this process involves numerous technical challenges. Future evaluation methodologies should place greater emphasis on assessing a model's ability to integrate information across modalities. Benchmarks that specifically test cross-modal reasoning, such as understanding the relationship between visual scenes and accompanying dialogue or background music, would provide valuable insights into VideoLLMs' multimodal capabilities. Additionally, metrics designed to evaluate the alignment and coherence of information across modalities would enhance the comprehensiveness of VideoLLM evaluation.

Developing benchmarks with controlled multimodal complexity would be particularly valuable, allowing for systematic evaluation of how models handle increasing levels of cross-modal integration. This could involve variations of the same content with different combinations of modalities present (e.g., video-only, video+audio, video+text, and full multimodal integration), enabling assessment of how each additional modality contributes to understanding. Another promising direction is the development of evaluation methodologies that assess models' ability to identify and resolve conflicts or ambiguities between modalities. In real-world videos, visual content may sometimes contradict audio narratives or text descriptions, and sophisticated models should be able to recognize and reason about such discrepancies.

\subsection{Addressing Hallucination in VideoLLMs}

``Hallucination" refers to the phenomenon where models generate responses that are significantly disconnected from the relevant source material or input, leading to highly erroneous or unrealistic descriptions that do not align with the provided videos. This issue poses significant challenges for the reliable evaluation of VideoLLMs. In video understanding with LLMs, hallucinations can arise from several sources: insufficient extraction of visual features, the influence of video contextual content leading to misinterpretations, the domain gap between visual feature representation and language representation, and the inherent tendency of LLMs to generate plausible but incorrect information. Future evaluation methodologies should incorporate specific measures to assess hallucination in VideoLLMs. This could include factual consistency checks that compare model outputs with ground-truth information, adversarial examples designed to trigger hallucinations, and metrics that specifically quantify the degree of hallucination in model responses. Additionally, addressing hallucination will require improvements in video encoders, enhanced understanding of long-form spatiotemporal contexts, and better alignment between visual and linguistic latent spaces.

A particularly promising approach would be the development of hallucination-focused benchmarks that systematically test models on videos with increasing levels of ambiguity, complexity, or unusual content. Such benchmarks could help identify the conditions under which models are most prone to hallucination and guide the development of more robust architectures. Another valuable direction is the creation of evaluation methodologies that assess a model's ability to express uncertainty or acknowledge knowledge limitations rather than generating incorrect information. Models that can accurately recognize when they lack sufficient information to answer confidently represent an important advancement in reducing harmful hallucinations.

\subsection{Interpretability and Explainability}

As VideoLLMs become more complex and are deployed in critical applications, understanding how they arrive at their conclusions becomes increasingly important. Current evaluation methodologies primarily focus on performance outcomes rather than the reasoning processes behind those outcomes. Future evaluation frameworks should incorporate metrics for interpretability and explainability, assessing not just what the model outputs but how it arrives at those outputs. This could involve evaluating the quality of explanations provided by models, the transparency of reasoning processes, and the alignment of internal representations with human-understandable concepts.

Benchmarks specifically designed to test explainability could include tasks that require models to justify their answers with evidence from the video, identify the key frames or regions that influenced their reasoning, or demonstrate step-by-step reasoning processes. Such benchmarks would not only evaluate the correctness of responses but also the soundness and transparency of the reasoning behind them. Additionally, developing evaluation methods that can assess the internal representations and attention patterns of VideoLLMs would provide valuable insights into how these models process and integrate information across frames and modalities. This could involve analyzing attention maps, feature activations, or other internal metrics to understand what aspects of the video the model focuses on when answering different types of questions.

\subsection{Robustness and Generalization}

Current evaluation methodologies often assess VideoLLMs on curated benchmarks that may not fully represent the diversity and complexity of real-world videos. This raises questions about the robustness of these models to variations in video quality, content, and context, as well as their ability to generalize across different domains and applications. Future evaluation frameworks should place greater emphasis on assessing robustness and generalization capabilities. This could involve testing models on videos with various levels of quality degradation (e.g., low resolution, poor lighting, camera shake), diverse visual styles and production qualities, and content from domains not represented in training data.

Developing domain-specific benchmarks would be particularly valuable for assessing generalization across different fields such as healthcare, education, surveillance, and entertainment. Each domain presents unique challenges and requirements, and a comprehensive evaluation should include an assessment of how well models can adapt to these diverse contexts. Another important direction is the creation of adversarial benchmarks designed to identify failure modes and limitations of current models. These could include videos designed to challenge common assumptions or exploit known weaknesses in video understanding systems, providing insights into areas needing improvement.

\subsection{Ethical and Responsible Evaluation Frameworks}
As VideoLLMs become increasingly powerful and widely deployed, ethical considerations in their evaluation become paramount. Current evaluation methods often focus primarily on performance metrics without sufficient attention to potential biases, fairness issues, or societal impacts. Future evaluation frameworks should incorporate specific metrics and tests for bias detection across different demographics, cultures, and content types. This could include measuring performance disparities across videos featuring different groups of people or assessing whether models reinforce harmful stereotypes in their interpretations.
Additionally, privacy concerns should be addressed in evaluation methodologies, particularly for models that may identify or track individuals in videos. Benchmarks that specifically test a model's adherence to privacy standards and its ability to appropriately anonymize sensitive information when needed would provide valuable insights into responsible deployment. Establishing standardized ethical guidelines for VideoLLM evaluation, including transparency in reporting limitations and potential risks, would contribute to more responsible development and application of these technologies.

\subsection{Computational Efficiency Assessment}
The computational resources required for VideoLLM inference represent a significant practical constraint, particularly for long-form video analysis or real-time applications. Current evaluation methodologies rarely consider efficiency metrics as a formal part of model assessment. Future frameworks should incorporate standardized measurements of computational efficiency, including inference time, memory usage, and energy consumption across different video lengths and complexities.
Benchmarks that specifically evaluate the trade-offs between model performance and efficiency would provide practitioners with valuable information for selecting appropriate models for specific use cases. This could involve defining efficiency-normalized performance metrics that reward models achieving high accuracy with lower computational costs. Moreover, evaluation of model performance under resource constraints, such as limited memory or processing power, would better reflect real-world deployment scenarios, particularly for edge devices or consumer applications.

\subsection{Cultural and Contextual Understanding}
VideoLLMs must operate effectively across diverse cultural contexts, interpreting culturally specific visual cues, gestures, symbols, and scenarios. Current evaluation methods often utilize datasets with limited cultural diversity, potentially leading to models that perform well only within specific cultural contexts. Future evaluation frameworks should incorporate cross-cultural assessment, testing models on videos from various geographical regions, cultural traditions, and social contexts.
Benchmarks specifically designed to evaluate cultural awareness could include tasks requiring interpretation of culturally specific scenarios, recognition of diverse cultural references, and appropriate handling of culturally sensitive content. This would help identify limitations in models' ability to generalize across cultures and guide development toward more culturally inclusive systems. Additionally, evaluation methodologies should assess models' awareness of their own cultural limitations, rewarding systems that appropriately express uncertainty when faced with unfamiliar cultural contexts rather than making potentially inappropriate assumptions.

\subsection{Temporal Evolution and Concept Drift}
The world captured in videos constantly evolves, with changes in visual styles, popular culture references, events, and technologies. VideoLLMs must adapt to these changes to maintain relevance and accuracy over time. Current evaluation methodologies rarely assess how well models handle temporal evolution or concept drift. Future frameworks should incorporate longitudinal evaluation, testing models on videos from different time periods to assess their temporal generalization capabilities.
Benchmarks designed to evaluate temporal robustness could include tasks requiring interpretation of videos featuring outdated technologies, historical events, or changing cultural norms. This would help identify limitations in models' ability to contextualize content within appropriate time periods. Additionally, evaluation methodologies should assess how effectively models can be updated to incorporate new concepts, events, or cultural references without compromising performance on existing knowledge, a crucial capability for maintaining relevance in rapidly changing domains.

\subsection{Personalization and Adaptability Assessment}
Different users have varying needs, preferences, and contexts when interacting with VideoLLMs. Current evaluation methodologies typically assess models on standardized tasks without considering personalization capabilities. Future frameworks should incorporate metrics for evaluating how well models can adapt to individual user preferences, learning styles, and specific use cases.
Benchmarks designed to test personalization could include scenarios requiring adaptation to user feedback over multiple interactions, customization of responses based on user expertise levels, or tailoring of content summaries to specific user interests. Evaluation methodologies should reward models that effectively balance general knowledge with personalized insights, and that demonstrate improvement in user-specific performance over time. This dimension of evaluation is particularly important for educational applications, content recommendation systems, and assistive technologies where adaptation to individual needs is crucial.
\subsection{Interdisciplinary Evaluation Approaches}
VideoLLMs operate at the intersection of multiple disciplines, including computer vision, natural language processing, cognitive science, and domain-specific knowledge areas. Current evaluation methods often focus narrowly on computer vision or NLP metrics without incorporating insights from other relevant disciplines. Future evaluation frameworks should adopt interdisciplinary approaches, drawing on evaluation methodologies from fields such as cognitive psychology, education, human-computer interaction, and domain-specific areas like medicine, law, or creative arts.
Collaborations between AI researchers and experts from these diverse fields could yield novel evaluation methodologies that better capture the multifaceted nature of video understanding. For example, incorporating cognitive science principles could lead to benchmarks that assess how well model interpretations align with human cognitive processes. Similarly, involving domain experts in evaluation could ensure that models meet the specific requirements of specialized applications, such as medical video analysis or legal evidence interpretation.

\section{Future Benchmark Design}

Based on the challenges and limitations identified in current evaluation methodologies, we propose several directions for the development of next-generation VideoLLM benchmarks. These proposed benchmarks would address existing gaps in evaluation and provide more comprehensive assessment of model capabilities.

\subsection{Hierarchical Understanding Benchmark}

A hierarchical understanding benchmark would assess video comprehension at multiple levels of granularity within the same content. This benchmark would include a diverse collection of videos, each accompanied by questions that progressively probe deeper levels of understanding: (i) Basic recognition: Identifying objects, actions, and scenes. (ii) Contextual understanding: Interpreting relationships between elements. (iii) Temporal reasoning: Understanding sequences, causality, and changes over time. (iv) Abstract comprehension: Inferring intentions, emotions, and themes. (v) Counterfactual reasoning: Considering alternative scenarios and outcomes.

By evaluating performance across these hierarchical levels, this benchmark would provide insights into the depth of a model's understanding capabilities and identify specific areas for improvement. It would be particularly valuable for assessing fine-grained understanding while maintaining context within the broader narrative.

\subsection{Multimodal Integration Benchmark}

A multimodal integration benchmark would systematically evaluate how models leverage information across visual, audio, and textual modalities. This benchmark would include: (i) Modality correspondence: Assessing alignment between information in different modalities. (ii) Cross-modal inference: Requiring reasoning that integrates cues from multiple modalities. (iii) Modality conflict resolution: Identifying and resolving contradictions between modalities. (iv) Modality importance weighting: Determining which modality provides the most relevant information for different query types.
(v) Missing modality compensation: Evaluating performance when certain modalities are absent or degraded.

This benchmark would feature videos with carefully controlled multimodal complexity, allowing for systematic evaluation of how different modalities contribute to understanding. It would be particularly valuable for applications requiring integration of diverse information sources, such as multimedia analysis, surveillance, and assistive technologies.

\subsection{Long-form Narrative Benchmark}

A long-form narrative benchmark would focus specifically on evaluating models' ability to understand extended video content, tracking narrative elements across time. This benchmark would include: (i) Local-global coherence: Assessing understanding of both local scenes and global narrative.
(ii) Character tracking: Following individual characters across scenes and understanding their development.
(iii) Thematic understanding: Identifying recurring themes, motifs, and narrative arcs. (iv) Memory and recall: Testing ability to reference earlier content when answering later questions. (v) Summarization at multiple levels: Evaluating the generation of concise summaries at different granularities.

This benchmark would feature extended videos from diverse genres, including documentaries, movies, TV episodes, and instructional content. It would be particularly valuable for applications involving long-form content analysis, such as media cataloging, educational content development, and narrative understanding.

\subsection{Interactive Evaluation Benchmark}

An interactive evaluation benchmark would simulate real-world use cases by allowing for dynamic, conversational interaction between users and models. This benchmark would include: (i) Clarification seeking: Evaluating when and how models request additional information.
(ii) Response refinement: Assessing adaptation based on user feedback.
(iii) Follow-up handling: Testing coherent responses to follow-up questions.
(iv) Multi-turn reasoning: Evaluating extended conversations about video content.
(v) User guidance interpretation: Assessing response to user directions (e.g., ``focus on the left side").

This benchmark would move beyond static question-answering to evaluate models in more realistic interactive contexts. It would be particularly valuable for applications involving human-AI collaboration, such as creative content analysis, educational systems, and assistive technologies.

\subsection{Robustness and Adversarial Benchmark}

A robustness and adversarial benchmark would systematically evaluate model performance under challenging conditions. This benchmark would include: (i) Quality variations: Testing performance on videos with different resolutions, framerates, and compression artifacts. (ii) Visual perturbations: Evaluating robustness to lighting changes, occlusions, and camera movements. (iii) Ambiguous content: Assessing performance on videos with unclear or ambiguous elements. (iv) Out-of-distribution scenarios: Testing generalization to unusual or novel content. (v) Adversarial examples: Including videos specifically designed to challenge model assumptions.

This benchmark would help identify failure modes and limitations of current models, providing insights into their practical reliability in real-world settings. It would be particularly valuable for applications requiring high levels of robustness, such as autonomous systems, security applications, and critical decision support.

\subsection{Explainability Benchmark}

An explainability benchmark would focus on evaluating not just the correctness of responses but also the quality and transparency of the reasoning behind them. This benchmark would include: (ii) Evidence grounding: Assessing the ability to justify answers with specific video evidence. (ii) Reasoning transparency: Evaluating clarity and coherence of explanations. (iii) Uncertainty expression: Testing appropriate acknowledgment of limitations and uncertainties. (iv) Visual grounding: Requiring identification of specific regions or frames supporting conclusions. (v). Counterfactual explanation: Evaluating ability to explain why alternative interpretations are incorrect.

This benchmark would provide insights into the interpretability of model reasoning processes, which is increasingly important for applications involving critical decisions or collaborative analysis. It would be particularly valuable for domains such as healthcare, legal analysis, and educational applications, where explanation quality is as important as answer correctness.

\section{Applications of VideoLLMs}

VideoLLMs demonstrate remarkable versatility across diverse domains. Understanding these applications provides crucial context for developing robust evaluation methodologies and benchmarks.

\subsection{Multimedia Content Analysis and Distribution}

VideoLLMs now enable powerful multimedia content analysis capabilities previously unattainable with traditional systems. These models enhance search algorithms by understanding semantic content within videos \cite{mao2023large}, generate context-aware recommendations based on visual narrative elements \cite{ju2022prompting}, and automate subtitle generation across languages \cite{yang2023vid2seq}. VideoLLMs excel at precise content retrieval by analyzing visual semantics \cite{zhao2023lavila, jin2023text, jin2023diffusionret}, enabling users to locate specific moments within extensive video libraries. These models generate concise video summaries by identifying key visual and auditory elements \cite{Pramanick_2023_ICCV}, drastically reducing the time required for content review in news production, education, and entertainment. In professional workflows, VideoLLMs automate metadata tagging, streamline editing decisions, and power intelligent content discovery systems. With each advancement, these models transform how media content is created, distributed, and consumed, delivering increasingly personalized viewing experiences.

\subsection{Interactive Education and Accessibility}

VideoLLMs transform educational videos into interactive learning experiences by analyzing instructional content and generating appropriate scaffolding for learners \cite{gan2023large}. These models bridge accessibility gaps by translating sign language into text or speech \cite{liu2023survey, de2023machine}, making video content available to deaf and hard-of-hearing communities without manual interpretation. In gaming and interactive entertainment, VideoLLMs generate contextually appropriate dialogue, storylines, and procedural content \cite{mishra2023generating, koomen2023text}, creating more dynamic and responsive user experiences. Extended reality applications leverage these models to generate immersive narrative elements \cite{gokce2023role, jung2023xr, yu2024promptfix, huang2023egocentric}, while interface systems utilize video analysis to understand user context and provide situational assistance \cite{bi2023misar}. In robotics, models like SayPlan \cite{rana2023sayplan} enable autonomous systems to interpret visual environments and navigate complex spaces through the integration of 3D scene understanding with language models.

\subsection{Healthcare and Safety Systems}

VideoLLMs process medical imaging data and literature to support diagnostic workflows \cite{eysenbach2023role, liu2022beat, liu2022disco, liu2024emage}, identify relevant case studies, and generate patient-friendly explanations of medical concepts \cite{li2023llava}. Medical education applications include analyzing surgical videos to identify techniques and generate procedural documentation automatically. Security systems employ VideoLLMs to detect communication patterns indicating potential threats \cite{al2023chatgpt, mouratidis2023modelling}, identify anomalous behaviors in surveillance footage \cite{de2023socratic}, and process complex data streams for pattern recognition \cite{lee2023lanobert, almodovar2023logfit}. Cybersecurity applications include visual phishing detection and enhanced digital forensics \cite{tang2023graphgpt}. VideoLLMs enhance autonomous vehicle systems by enabling natural language interfaces \cite{cui2024drive}, interpreting traffic signage and road conditions \cite{li2023otter, lai2023lisa}, and improving situational awareness through multimodal understanding of the driving environment.

\subsection{Retail Analytics and Customer Experience}

VideoLLMs analyze in-store camera feeds to map customer journeys, identify engagement patterns with merchandise, and optimize store layouts based on behavioral analysis \cite{einfochips2025, staqu2025,electronosolutions2018}. These models enable visual search functionality where customers can submit images or short videos to find similar products, eliminating the need for precise textual descriptions. 
In e-commerce, VideoLLMs extract product features from demonstration videos, generate comparative analyses, and personalize product recommendations based on visual preference patterns. Virtual try-on systems leverage these models to realistically render products in personalized contexts, showing how items would appear in use. Customer service applications include analyzing customer-submitted videos of product issues to diagnose problems and recommend solutions without human intervention. By processing visual data at scale, VideoLLMs provide retailers with unprecedented insights into consumer behavior, allowing for more responsive inventory management and trend forecasting.

\section{Conclusion}

This survey has provided a comprehensive analysis of benchmarks and evaluation methodologies for VideoLLMs, highlighting their characteristics, strengths, limitations, and future directions. We have examined the evolution from closed-set evaluations with predefined answers to more flexible open-set approaches that leverage LLMs for scoring, as well as specialized methods for assessing temporal and spatiotemporal understanding. The landscape of VideoLLM benchmarks continues to evolve, with newer benchmarks incorporating longer videos, more diverse content, and more complex evaluation criteria. However, significant challenges remain in evaluating fine-grained understanding, long-form video comprehension, multimodal integration, human interaction, and hallucination mitigation. Based on these challenges, we have proposed several directions for next-generation benchmark design, including hierarchical understanding benchmarks, multimodal integration benchmarks, long-form narrative benchmarks, interactive evaluation benchmarks, robustness and adversarial benchmarks, and explainability benchmarks. These proposed benchmarks would address existing gaps in evaluation and provide a more comprehensive assessment of VideoLLM capabilities across multiple dimensions. The applications of VideoLLMs across media, interactive technologies, healthcare, and security underscore the importance of robust evaluation methodologies. As these models continue to evolve and find new applications, comprehensive and reliable evaluation will be essential for ensuring their effectiveness and reliability. By addressing the challenges and pursuing the future directions outlined in this survey, researchers can enhance the development and evaluation of VideoLLMs, ultimately leading to more capable, trustworthy, and useful systems for video understanding with large language models. 

{
   \small
   \bibliographystyle{IEEEtran}
   \bibliography{main}
}

\end{document}